\documentclass[runningheads]{llncs}
\usepackage{graphicx}

\usepackage{multirow}

\usepackage{textcomp}
\usepackage[toc]{glossaries}

\newacronym{cnn}{CNN}{Convolutional Neural Networks}
\newacronym{dcnn}{D-CNN}{Deep Convolutional Neural Networks}
\newacronym{froc}{FROC}{Free Receiver Operating Characteristic}
\newacronym{auc}{AUC}{Area Under the ROC Curve}

\newacronym{roc}{ROC}{Receiver Operating Characteristic}

\newacronym{mlo}{MLO}{Medio Lateral Oblique}
\newacronym{cc}{CC}{Cranio-Caudal}
\newacronym{cad}{CAD}{Computer-Aided Detection}
\newacronym{dm}{DM}{Digital Mammogram}
\newacronym{gap}{GAP}{Global Average Pooling}
\newacronym{fpr}{FPR}{False Positive Rate}
\newacronym{tpr}{TPR}{True Positive Rate}
\newacronym{fp}{FP}{False Positive}
\newacronym{tp}{TP}{True Positive}
\newacronym{dl}{DL}{Deep Learning}
\newacronym{cpm}{CPM}{Competition Performance Metric}

\newacronym{pauc}{PAUC}{Partial Area Under the FROC Curve}

\newacronym{sgd}{SGD}{Stochastic Gradient Decent}
\newacronym{roi}{ROI}{Region of Interest}

\makeglossaries
\begin{document}
\title{Improving Breast Cancer Detection using Symmetry Information with Deep Learning} 
\titlerunning{Improving Breast Cancer Detection using Symmetry Information}
%
\author{Yeman Brhane Hagos\inst{1,3} \and
Albert Gubern M\'erida\inst{1}\and
Jonas Teuwen\inst{1,2}}


\institute{Radboud University Medical Center, Department of Radiology and Nuclear Medicine,
Nijmegen, the Netherlands \and
Delft University of Technology, the Netherlands \and 
University of Burgundy (France), University of Cassino and Southern Lazio (Italy) and University of Girona (Spain)
}
\maketitle              
%
\begin{abstract}
\gls{cnn} have had a huge success in many areas of computer vision and medical image analysis. However, there is still an immense potential for performance improvement in mammogram breast cancer detection \gls{cad} systems  by integrating all the information that radiologist utilizes, such as symmetry and temporal data. In this work, we proposed a patch based multi-input \gls{cnn} that learns symmetrical difference to detect breast masses.  The network was trained on a large-scale dataset of $28294$ mammogram images. The performance was compared to a baseline architecture without symmetry context using \gls{auc} and \gls{cpm}. At candidate level, \gls{auc} value of $0.933$ with $95\%$ confidence interval of $[0.920\  ,\  0.954]$ was obtained when symmetry information is incorporated in comparison with baseline architecture which yielded \gls{auc} value of $0.929$ with $[0.919\  ,\  0.947]$ confidence interval. By incorporating symmetrical information, although there was no a significant candidate level performance again ($p = 0.111$), we have found a compelling result at exam level with \gls{cpm} value of $0.733$ ($p = 0.001$). We believe that including temporal data, and adding benign class to the dataset could improve the detection performance.
\end{abstract}
\section{Introduction}
Breast cancer is the second most common cause of cancer death in women after lung cancer in the United States, which covers around $30\%$ of cancers diagnosed and the chance of women dying from breast cancer is $2.6\%$ ~\cite{rakhlin2018deep}. Mammography is the main imaging modality used to detect breast abnormalities at an early stage. Breast masses are most dense and appear in grey to white pixel intensity with oval or irregular shape~\cite{oliver2010review}. Normally, irregular shaped masses are suspicious \cite{{oliver2010review}}, \cite{dhungel2017fully}. Breast cancer screening has shown a reduction in mortality rate of between $40\%$ and $45\%$  for women who were undergoing mammogram screening regularly ~\cite{feig2002effect}. However, mammogram screening has drawbacks due to \gls{fp} recalls, such as \gls{fp} biopsy and cost associated with the unnecessary follow up~\cite{geras2017high}. Therefore, it is necessary to increase sensitivity for early stage detection and increase specificity to reduce \gls{fp} detection.
 
Nowadays, with a massive amount of data and computational power, \gls{dl} has shown a remarkable success in  natural language processing ~\cite{bahdanau2014neural} and object detection and recognition~\cite{wang2016deep}. This has opened an interest in applying \gls{dl} in medical image processing and analysis. However, care should be taken as the way we as humans interpret natural images and medical images are different in some cases. Eventually, the performance of \gls{dl} method will be compared with the radiologist and thus, the \gls{cnn} should preferably be given all the information that radiologist utilize. For instance, during the reading of screening mammograms, radiologists use priors, multiple views and look for asymmetries between the two breasts.

\gls{dl} has been explored for \gls{dm} image analysis. Some of them work directly on the whole image \cite{geras2017high}, \cite{dhungel2017fully}, and  others focused on patch based \cite{kooi2017large}. \cite{geras2017high} proposed a multi-view single stage \gls{cnn} breast mammogram classification that works at original resolution. To address memory issue, aggressive convolution and pooling layers with stride greater than one were proposed. It is stated also that this approach suffers from loss of spatial information. In the work by \cite{kooi2017large}, incorporating symmetry and temporal context improves detection of malignant soft tissue lesion, in which random forest classifier was used for mass detection and \gls{cnn} for classification.

In this study, we conducted an investigation to analyze the performance gain of integrating symmetry information into a \gls{cnn} to detect malignant lesions on a large scale mammography database. First, a database of $7196$ exams which contains 28294 images was collected from different sites in the Netherlands. Previous work by \cite{karssemeijer1999local} was employed to detect suspicious candidates locations. Then, patches centered on the points were extracted to train a two input \gls{cnn} to reduce \gls{fp} candidates. Left and right breast images were considered as contra-later images to each other, and a patch in a primary image and an exact reflection or mirror on the contra-lateral were considered as a pair of inputs to the network.
\section{Materials and Methods}
\subsection{Dataset}
The mammogram images used were collected from General Electric, Siemens, and Hologic from women attending for diagnostic purpose between $2000$ and $2016$. The images are anonymized and approved by the regional ethics board after summary review, with a waiver of a full review and informed consent ~\cite{de2018automated}. The database contains $7196$ exams. For most of the exams, \gls{mlo} and \gls{cc} views of both  breasts are provided, resulting in $28294$ \gls{dm} images in totals. All images with malignant lesions were histopathologically confirmed, while normal exams were selected if they had at least two years of negative follow-up. From $7196$ \gls{dm} exams, $2883$ exams ($42\%$) contained a total of $3023$ biopsy-verified malignant lesions. The exact distribution of the dataset is shown in Table \ref{tab:data}. In the whole dataset, $1315$ exams does not have either left or right breast images of \gls{mlo} and/or \gls{cc} views.

Training, validation and test data split  was  done  at  patient level  to evaluate the generalization of the the model developed. Data was randomly split into training ($50\%$), validation ($10\%$) and testing ($40\%$) while making sure exams from each vendors present in each partition proportionally.
\begin{table*}
\centering
\caption{Distribution of \gls{dm} dataset used including their vendor.}\label{tab:data}
\begin{tabular}{|l|c|c|c|}
\hline
 &  General Electric     & Siemens & Hologic\\
\hline
number of studies &  2248 & 1518 & 3430\\
normal images    &  7771 & 5842 & 12288\\
images with malignant lesions    & 1292 & 255 & 1476\\
\hline
\end{tabular}
\end{table*}
\begin{figure}[h]
    \centering
    \begin{minipage}[b]{0.6\linewidth}
        \includegraphics[width=\textwidth]{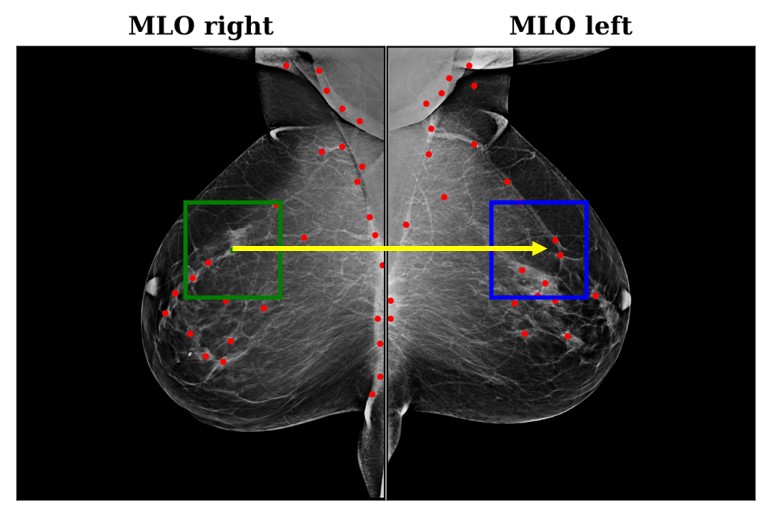}
    \end{minipage}
    \caption[]{An illustrative example showing center location of suspected masses in a sample \gls{mlo} views of right and left breast mammogram images, and patches used to train symmetry \gls{cnn} model. The green box represents a patch centered on a positive candidate on \gls{mlo} view of right breast \gls{dm} image, and its corresponding symmetry patch at the same location on the contra-lateral image is displayed in blue.}
    \label{fig:sample_images}
\end{figure}
\subsection{Candidate Selection} \label{sec:candidate_selcetion}
Previous work by \cite{karssemeijer1999local} was employed to detect suspicious mass candidates. Fig. \ref{fig:sample_images}) shows sample \gls{mlo} view \gls{dm}  images of left and right breast. Likelihood of a pixel to be part of a mass was computed using local lines and distribution of gradient orientation features. Then, a global threshold was applied to the likelihood image to generate regions that are considered as suspicious. The red and green points correspond to suspected candidate center locations of mass. The green point is a true mass and others are false positive candidates. Table \ref{tab:cand_data} details the number of suspicious candidates from training, validation and test data.
\begin{table}[h]
\centering
\caption{Number of suspicious candidates. The numbers after + indicate candidates from exams  without left or right breast images. Positive refers to candidates inside malignant mass and negative candidates are outside a mass.}\label{tab:cand_data}
\begin{tabular}{|l|c|c|c|}
\hline
 Candidates &  Training     & Validation & Test\\
\hline
 negative &  337366+2359 & 61833 +1093 & 250293+6154\\
 positive    &  2217+58 & 927+30 & 727+67\\
\hline
\end{tabular}
\end{table}
\section{Patch Extraction and Augmentation} \label{sec:patchpre}
To extract patches, the contra-lateral images were flipped horizontally to place both images in the same space. Maximum size of the mass in our dataset was about $5 \text{cm}$, and a patch size of $300 \times 300$ pixels ($6\text{cm} \times 6 \text{cm}$) was considered to provide enough context to discriminate soft tissue lesions. We reduced the number of training negative candidates by ensuring a sufficient distance (at least $2 \text{cm}$) from a lesion and an inter-negative candidate distance of $1.4 \text{cm}$. This resulted in $253476$ ($74.6\%$) negatives patches.

As an augmentation scheme, initially positive patches were flipped and Gaussian blurred with standard deviation between $[0.2, 3]$. Then, with probability of $0.5$ one of the three augmentations were applied to both negative and positive patches: scaling, translation around the center and rotation. The parameters for these augmentations were uniformly selected from $[0.88 \ , 1.25]$, $[-25 \ , 25]$ and [$-30$\textdegree, $30$\textdegree], respectively. 
\section{Network Architecture and Training}
\begin{figure}
    \centering
    \begin{minipage}[b]{0.49\linewidth}
        \centering
        \includegraphics[width=\textwidth]{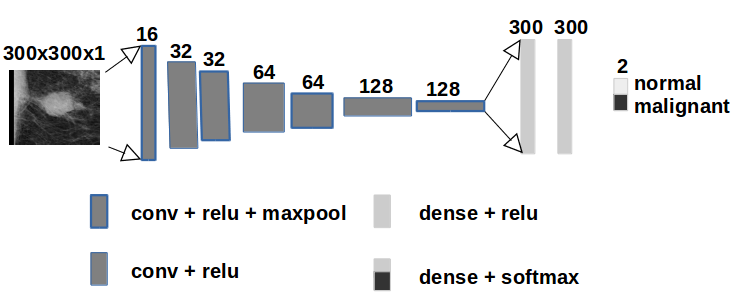}
        \centering
        \ a)
    \end{minipage}
    \hfill
    \begin{minipage}[b]{0.49\linewidth}
        \centering
        \includegraphics[width=\textwidth]{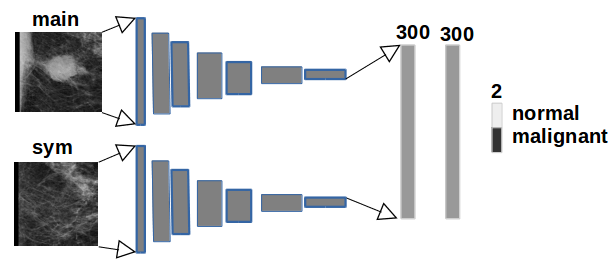}
        \centering
        \ b)
    \end{minipage}
    \caption[]{\gls{cnn} architectures: a) Baseline architecture b) Symmetry model architecture}
    \label{fig:cnn_models}
\end{figure}

In addition to incorporating symmetry information, a single input  baseline architecture was trained. The baseline architecture is a variant of VGG architecture \cite{simonyan2014very} as shown in Fig. \ref{fig:cnn_models}a and it consists of feature extraction and classification parts. The feature extraction section has a series of seven convolutional layers with $\{16, 32, 32, 64, 64, 128, 128\}$ filters each followed by a max pooling layer. Convolution was performed with a stride of ($1,1$) and valid padding. The classification part is composed of three dense layers with $\{300, 300,2\}$ neurons and with dropout (rate= $0.5$) regularization after the first dense layer. ReLU activation was chosen for all layers but softmax for the last. \gls{gap}~\cite{lin2013network} was applied after the last convolutional layer while the other layers are followed by $3\times3$ max pooling. The advantage of \gls{gap} over flattening is it minimizes overfitting by reducing the number of parameters. The symmetry model has two inputs, the primary patch and a contra-lateral patch as shown in Fig. \ref{fig:cnn_models}b. The parallel streams were transfer learned from the baseline architecture in Fig. \ref{fig:cnn_models}a without weight sharing. The features from the parallel stream were concatenated and fed to the classifier. The classification part is similar to the baseline model. For exams without a contra-lateral image, zero matrices were used as a symmetry image.

Weights of both networks were initialized using Glorot weight initialization and optimized using \gls{sgd} with time-based learning rate scheduler with an initial learning rate (ILR) of $10^{-2}$ for baseline architecture and $ 10^{-3}$ for the symmetry model, decay rate(ILR/$200$), and momentum ($0.9$). Mini-batch size of $64$ was used and for each epoch, all positives samples were presented twice and an equal number of randomly sampled negatives, ensuring balance in each batch. Model with highest \gls{auc} on validation was selected as the best model. Furthermore, we monitored \gls{auc} for early stopping with patience of $20$ epochs. 
\section{Result} \label{sec:result}
All the experiments were conducted in Keras\cite{chollet2017keras}, and results presented are on a separately held $40\%$ of the data. Candidate level quantitative evaluation was done using \gls
{auc} and \gls{froc} along with \gls{cpm}~\cite{setio2017validation} for image and exam level performance analysis. Moreover, a $95\%$ confidence interval and p-values of \gls{auc} and \gls{cpm} were computed using bootstrapping \cite{efron1994introduction}, using $1000$ bootstraps.

\begin{figure}[h]
    \centering
    \begin{minipage}[b]{0.45\linewidth}
        \centering
        \includegraphics[width=\textwidth]{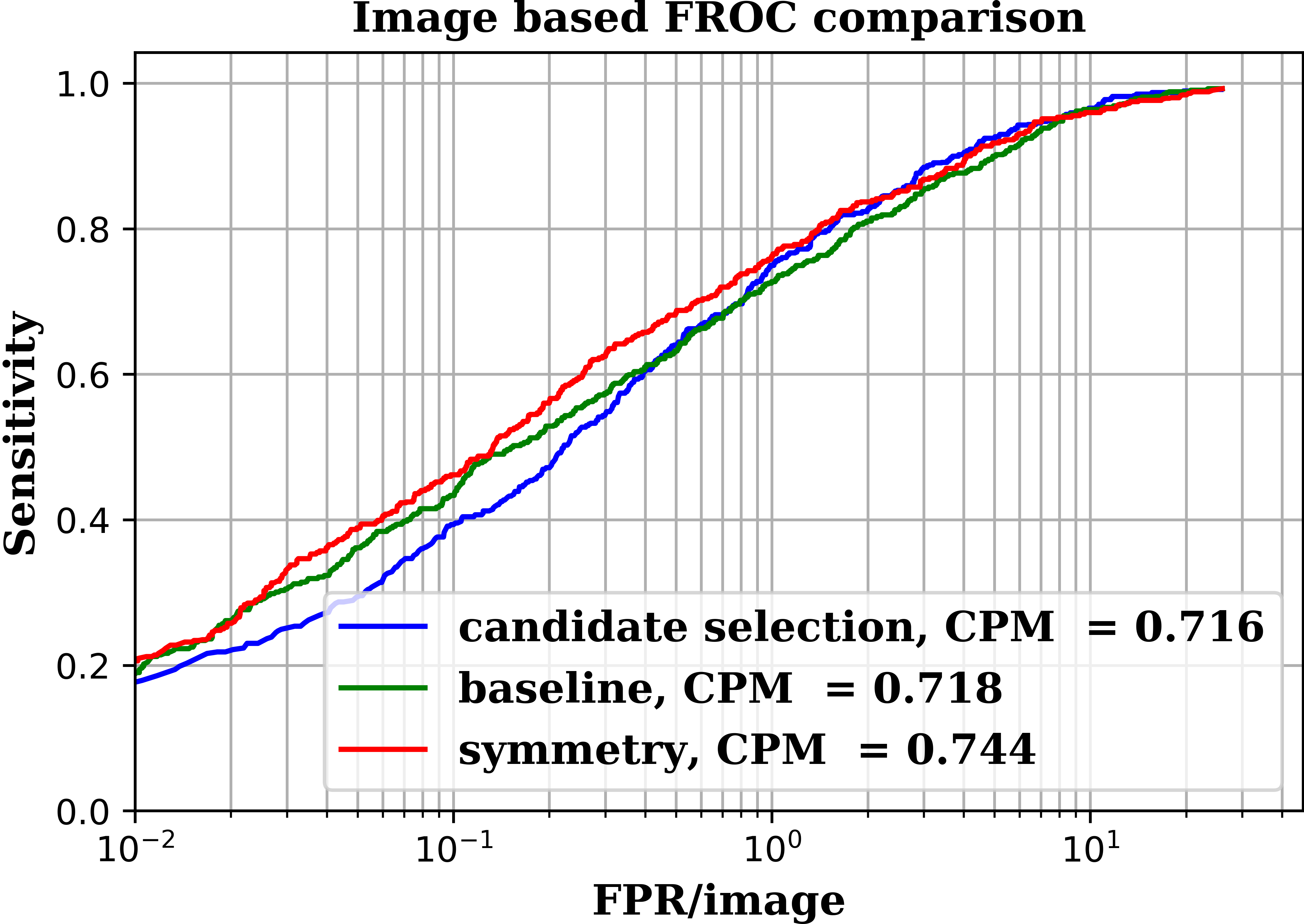}
        \centering
        \ a)
    \end{minipage}
    \begin{minipage}[b]{0.45\linewidth}
        \centering
        \includegraphics[width=\textwidth]{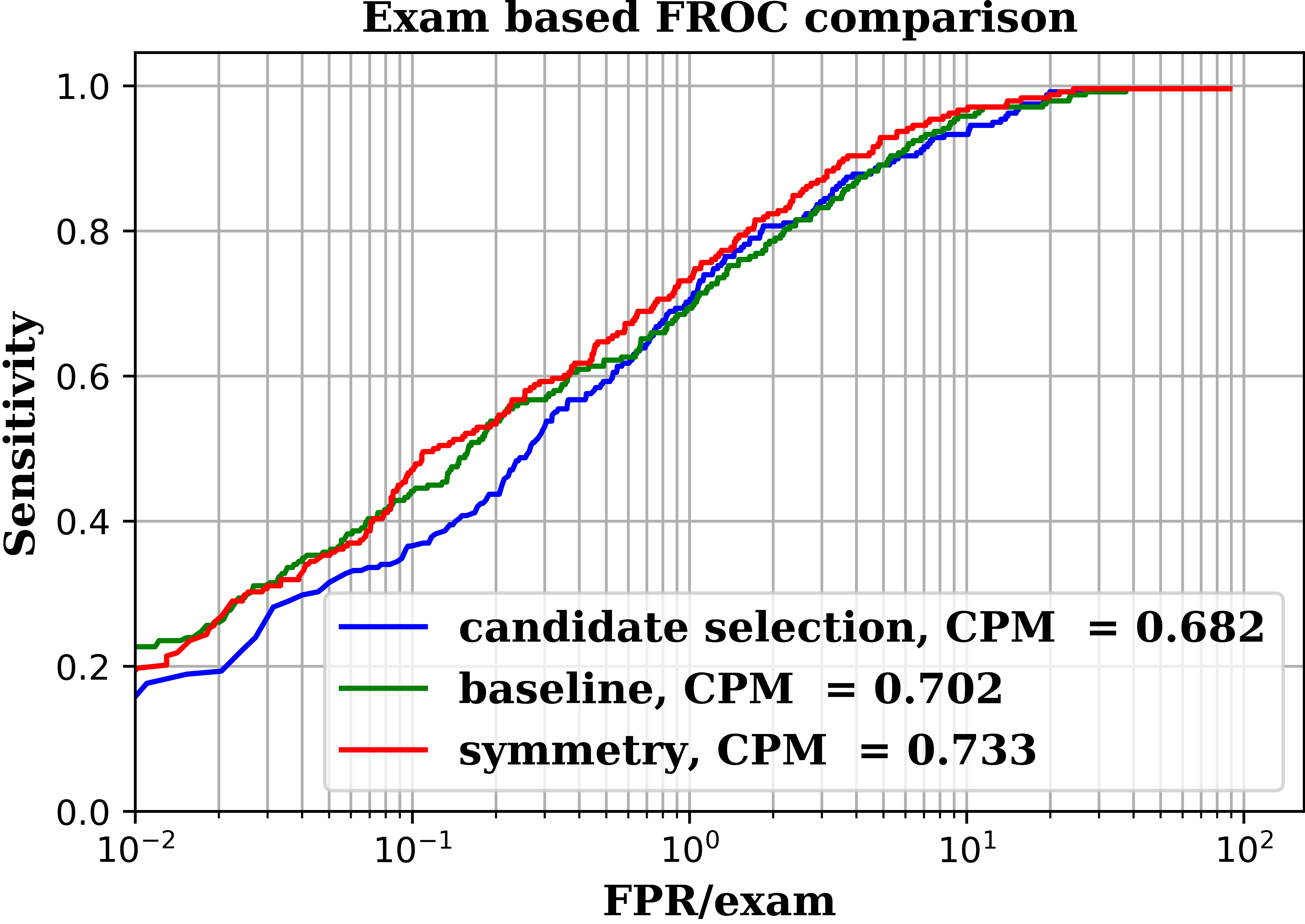}
        \centering
        \ b)
    \end{minipage}
\caption{ \gls{froc} comparison of candidate selection, baseline and symmetry models; a) Image based \gls{froc}.  b) Exam based \gls{froc}.}
    \label{fig:froc}
\end{figure}

Table \ref{tab:roc} reveals \gls{auc} values of the candidate selection, baseline and symmetry models. \gls{auc} value of $0.896$ with $95\%$ confidence interval of $[0.879\  ,\  0.913]$ was obtained by the model used for candidate selection. The baseline architecture that processes a single \gls{roi} image yielded an \gls{auc} value of $0.929$ with $95\%$ confidence interval $[0.916\ ,\  0.942]$, which is significantly better than the candidate selection stage performance ($p = 0.004$). Incorporating symmetry information improved the \gls{auc} to $0.933$ with $[0.919\  ,\  0.947]$ $95\%$ confidence interval, although it was not significant ($p = 0.111$) in comparison with baseline architecture. For symmetry model, a zero matrix was used as a substitute when contra-lateral image is missing and a separate evaluation is presented in Table \ref{tab:roc}. The separate evaluation resulted in an \gls{auc} value of $0.866$ with $95\%$ significance interval of  [$0.788 \ ,\  0.930$]. A symmetry model was trained without augmentation and the best model resulted in \gls{auc} value of $0.91$. This shows the proposed augmentation has significantly improved detection \gls{auc}.
\begin{table}
\centering
\caption[]{\gls{auc} comparison of candidate selection, baseline and symmetry network. symmetry$^*$ represents evaluation of symmetry model on candidates with missing contra-lateral patch.}\label{tab:roc}
\begin{tabular}{|l|c|c|c|c|}
\hline
    & candidate selection & baseline & symmetry & symmetry$^*$\\
\hline
\textbf{AUC} &  0.896 & 0.929 & 0.933 & 0.866 \\
\hline
\end{tabular}
\vspace*{-4mm}
\end{table}
Fig. \ref{fig:froc} present image and exam based \gls{froc} comparison of the three models. In our test set, the symmetry model showed a better performance ( $p = 0.001$) compared to the baseline architecture at both image and exam level. At an image level, \gls{cpm} value of 0.716, 0.718, and 0.744 with $95\%$ confidence interval of $[0.682\  ,\  0.750]$, $[0.679\  ,\  0.756]$ , and $[0.723\  ,\  0.794]$ was obtained for candidate selection, baseline and symmetry model, respectively. Moreover, during exam level evaluation sensitivity of the model that incorporates symmetry context was found to be better than the other model throughout the whole \gls{fpr} range, resulting in \gls{cpm} value and confidence interval of $0.733\ [0.721\ ,\  0.823]$ compared to $0.682\ [0.671\ ,\  0.746] $ and $0.702 \ [0.687 \ ,\  0.772]$ for candidate selection and baseline model, respectively. 
\section{Discussion}\label{sec:discussion}
\begin{figure}
    \centering
    \begin{minipage}[b]{0.3\linewidth}
        \centering
        \includegraphics[width=\textwidth]{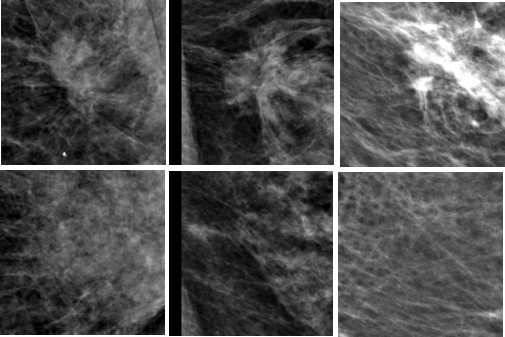}
    \centering
    \ a) 
    \end{minipage}
    \hspace{1em}
    \begin{minipage}[b]{0.3\linewidth}
        \centering
        \includegraphics[width=\textwidth]{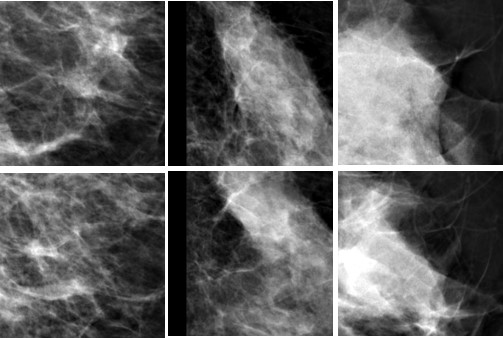}
    \centering
    \ b)
    \end{minipage}
    \caption[]{Sample patches with an improved prediction using symmetry model: a)  Positive patches that were misclassified by baseline architecture and correctly classified by symmetry model. b) Negative patches that were misclassified by baseline architecture and correctly classified by symmetry model. The top and bottom row images are primary and contra-lateral pairs, respectively.}
    \label{fig:improved}
\end{figure}
The proposed patch augmentation method showed an improvement in the generalization of the \gls{cnn} model and thus, the performance of the classifier. Symmetry model trained without patch augmentation yielded \gls{auc} value of $0.91$ on a test set, in comparison to $0.933$ when augmentation was applied. Moreover, incorporating symmetry information helps in learning distinctive features when there is a low-intensity contrast between mass and the background as shown in Fig. \ref{fig:improved}a. For the malignant candidates in Fig. \ref{fig:improved}a, without symmetry context malignancy probability was found to be below $0.2$, however, integrating symmetrical information increased the malignancy prediction to a value greater than $0.7$. Moreover, the negative patches in Fig. \ref{fig:improved}b were predicted as malignant masses by the baseline model (probability greater than $0.9$), however, after including symmetrical context, their malignancy probability was found less than $0.1$.

One of the main limitations of this work is that only soft tissue lesions were studied and detecting calcification will be of added value. Secondly, some benign abnormalities were found to be difficult for the network to differentiate from malignant candidates. We expect that separating the benign candidates from the normal and training with three classes could improve the detection performance. As studied in \cite{kooi2017large}, integrating temporal context could also improve the performance of the model.
\section{Conclusions}\label{sec:conclusions}
In this work, we proposed a deep learning approach that integrates symmetrical information to improve breast mass detection from mammogram images. Previous work by Karssemeijer et al. \cite{karssemeijer1999local} was used to detect suspicious candidates. The \gls{fp} candidates were reduced by learning symmetrical differences between primary and contra-lateral patches. \gls{auc} was employed as a performance measure at candidate level, whilst \gls{cpm} was computed for image and exam level evaluation. We have found that our proposed approach reduces \gls{fp} predictions compared to baseline architecture. An AUC value $0.933$ ( $p = 0.111$) with $95\%$ confidence interval of [$0.919 \ , \ 0.947$] was obtained at candidate level and $0.733$ ($p = 0.001$) \gls{cpm} with $95\%$ confidence interval of [$0.721\ ,\  0.823$] was achieved with our symmetry model.

Training with a dataset which includes more time points could possibly improve reliability and detection accuracy \cite{kooi2017large}, and will be part of our future work.

\end{document}